\documentclass{article}
\usepackage{ijcai25}
\usepackage{times}
\usepackage{soul}
\usepackage{url}
\usepackage[hidelinks]{hyperref}
\usepackage[utf8]{inputenc}
\usepackage[small]{caption}
\usepackage{graphicx}
\usepackage{amsthm}
\usepackage{booktabs}
\usepackage[switch]{lineno}
\usepackage{multirow}
\usepackage{algpseudocode}
\usepackage{amsmath}
\usepackage{amssymb}
\usepackage{enumitem}
\usepackage{xcolor}
\usepackage{hyperref}
\usepackage{booktabs}
\usepackage{caption}
\usepackage{amsfonts}

\title{PRE-MAP: Personalized Reinforced Eye-tracking Multimodal LLM for High-Resolution Multi-Attribute Point Prediction}

\author{
Hanbing Wu$^{1*}$ \quad 
Ping Jiang$^{2*}$ \quad 
Anyang Su$^{3*}$ \quad 
Chenxu Zhao$^3$ \quad 
Tianyu Fu$^3$ \\
Minghui Wu$^3$ \quad 
Beiping Tan$^3$ \quad 
Huiying Li$^{1\dagger}$ \\
\affiliations
$^1$Jilin University \quad
$^2$Peking University \quad
$^3$Mininglamp Technology \\
\emails
 wuhb23@mails.jlu.edu.cn, 
2101120066@stu.pku.edu.cn,
\{suanyang,zhaochenxu,futianyu,wuminghui,tanbeiping\}@mininglamp.com,
lihuiying@jlu.edu.cn
}

\begin{document}

\maketitle
\let\thefootnote\relax\footnotetext{
$^{*}$ Authors contributed equally to this work. \\
$^{\dagger}$ Corresponding authors.
}

\begin{abstract}
Visual selective attention, driven by individual preferences, regulates human prioritization of visual stimuli by bridging subjective cognitive mechanisms with objective visual elements, thereby steering the semantic interpretation and hierarchical processing of dynamic visual scenes. However, existing models and datasets predominantly neglect the influence of subjective cognitive diversity on fixation behavior. 
Conventional saliency prediction models, typically employing segmentation approaches, rely on low-resolution imagery to generate saliency heatmaps, subsequently upscaled to native resolutions, which limiting their capacity to capture personalized attention patterns. Furthermore, MLLMs are constrained by factors such as hallucinations, making it very costly to strictly adhere to the expected format in tasks involving multiple point predictions, and achieving precise point positioning is challenging. To address these limitations, we present \textbf{S}ubjective \textbf{P}ersonalized \textbf{A}ttention for \textbf{Ad}vertisement \textbf{V}ideos, namely \textbf{SPA-ADV}, a large-scale multimodal dataset capturing gaze behaviors from over 4,500 participants varying in age and gender with 486 videos.
Furthermore, we propose \textbf{PRE-MAP}, a novel eye-tracking saliency model that characterizes \textbf{P}ersonalized visual disparities through \textbf{R}einforcement learning-optimized \textbf{E}ye-tracking, built upon MLLMs and guided by \textbf{M}ulti-\textbf{A}ttribute user profiles to predict \textbf{P}oints. To ensure MLLMs produce prediction points that are both format-correct and spatially accurate, we introduce Consistency Group Relative Policy Optimization (\textbf{C-GRPO}), inspired by the variability in eye movement points and Multi-Attribute profiles. Extensive experiments on SPA-ADV and other benchmarks demonstrate the effectiveness of our approach. The code and dataset are available at \href{https://github.com/mininglamp-MLLM/PRE-MAP}{this URL}.
\end{abstract}

\begin{figure}[t]
  \centering
  \includegraphics[width=1\linewidth]{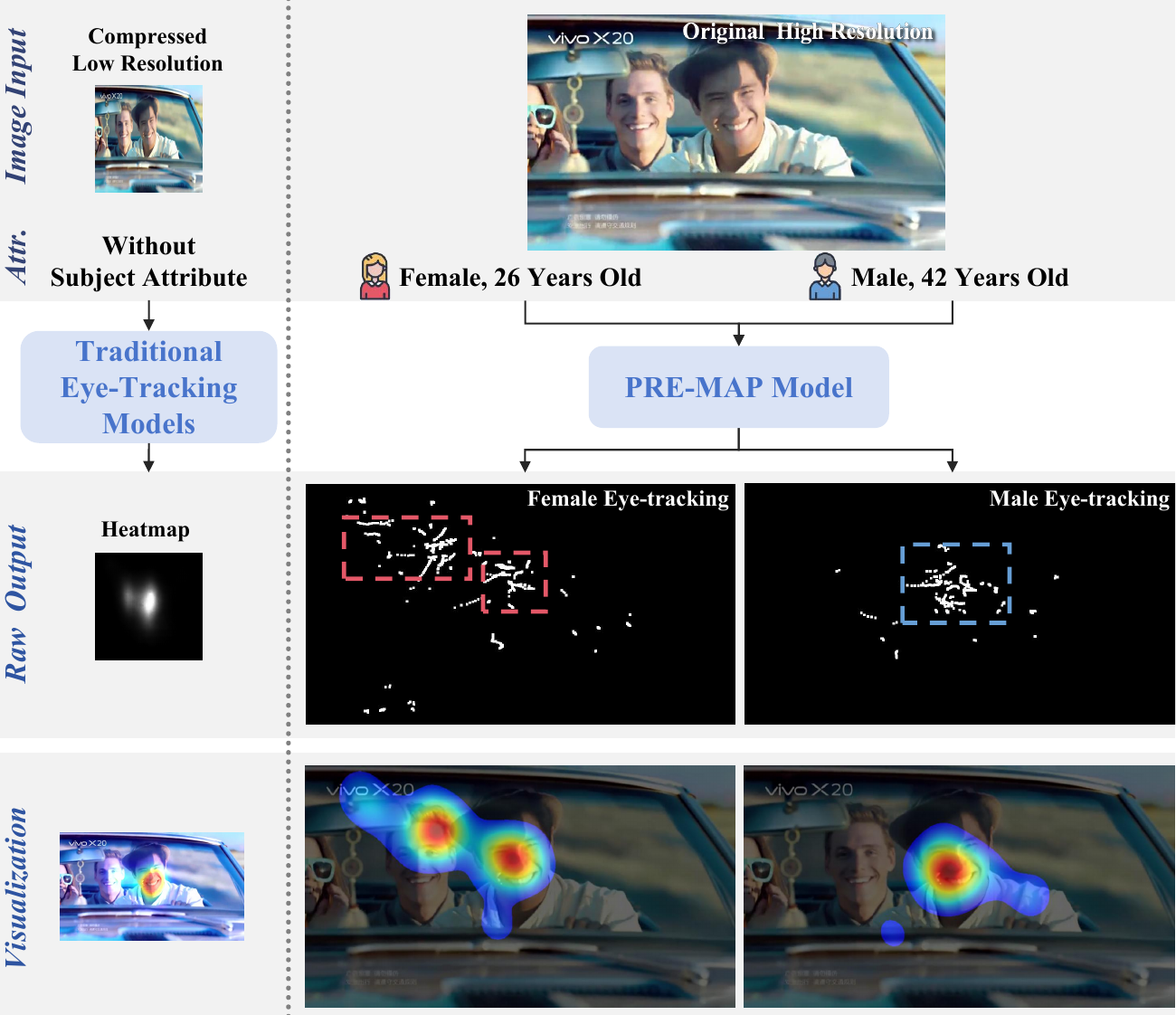}
  \caption{Traditonal eye-tracking models, typically limited by subject-independent designs and low-resolution input processing, produce population-averaged saliency maps that compromise fine-grained visual details, often leading to artifacts during image upscaling applications. Under identical visual stimuli, our model successfully captures gender-based differences in visual attention patterns. These simulated behavioral differences align with original empirical trends, as demonstrated in the right.}
  \label{introduction}
\end{figure}

\section{Introduction}
\label{sec:intro}
Modeling subjective human visual attention plays a crucial role in advancing fields like computer vision and cognitive science, with applications including advertisement optimization, human-computer interaction, and medical image analysis.~\cite{science,itro1,itro2,itro3,eeg}. Visual selective attention~\cite{selective} describes the human capacity to prioritize certain visual stimuli while ignoring others.  As shown in Figure~\ref{introduction}, when exposed to identical visual stimuli, subtle gender-related differences in attention deployment emerge~\cite{gender}: female participants exhibit broader gaze distributions and higher saccadic frequencies, whereas males show more localized and rapid fixation patterns. This process reflects the dynamic interaction between objective visual signals and subjective cognitive factors, including personal preferences, prior experiences, and emotional states \cite{social1,social2,social3}. However, effectively modeling such heterogeneous attention patterns remains a substantial challenge for current computational approaches.

Traditional saliency prediction models~\cite{SUM,tran} primarily employ computer vision paradigms and typically handle inputs of specific resolutions (typically below 512px), relying solely on visual modalities. These models generate population-level attention representations through statistical aggregation of multi-observer eye-tracking data. However, they often fail to capture the subjective cognitive diversity among individuals, facing several significant challenges: (1) The process of downsampling images before generating heatmaps and subsequently upscaling them results in pixelation and the loss of fine-grained attention details; (2) Personalization is frequently neglected or inadequately addressed, which undermines scalability and generalization; (3) Existing datasets tend to have limited participant samples, introducing subjective biases and constrains the potential for large-scale studies on attention variability across different individuals.

To advance research on selective visual attention influenced by subjective cognitive differences, we introduce \textbf{S}ubjective \textbf{P}ersonalized \textbf{A}ttention for \textbf{A}dvertisement \textbf{V}ideos (\textbf{SPA-ADV}). This large-scale eye-tracking collection captures individual gaze patterns under consistent visual stimuli. SPA-ADV provides precise fixation coordinates from participants representing diverse demographic and cognitive profiles, offering a comprehensive foundation for studying personalized saliency and inter-individual attention variability.

Inspired by personalized attention eye-tracking data from the SPA-ADV dataset, we propose a novel framework for subjective saliency prediction: the Multi-Attribute Point-Based Attention Model (\textbf{PRE-MAP}). Departing from conventional heatmap regression, we aim to achieve a more natural alignment with human gaze behavior through a fine-grained point prediction paradigm\cite{Scanpath,Scanpath1}. The PRE-MAP leveraging the multi-attribute textual knowledge and fine-grained visual localization capabilities of multimodal large language models (\textbf{MLLMs}). Our model integrates user-specific attributes (including age, gender, and cognitive style) with video content analysis to capture how subjective cognition influences attention distribution. It predicts personalized fixation points on one-second high-resolution video clips without relying on saliency segmentation modules or specialized vision decoders. Additionally, by incorporating step-wise supervised fine-tuning (SFT) and task-specific learnable tokens, we enhance the model's capability in modeling personalized visual attention.

Despite providing the model with foundational fixation prediction capabilities, SFT alone fails to ensure structural consistency and spatial accuracy in gaze prediction. This limitation stems primarily from the variable quantity of fixation points and the dynamic characteristics of output sequences, resulting in challenges such as coordinate count mismatches and spatial position drifting. The aforementioned issues can be effectively mitigated through reinforcement learning coupled with our specifically designed reward functions. Firstly, the format consistency reward is designed to enforce numerical consistency by aligning the number of predicted fixation points to eliminate quantity mismatches. Secondly, we implement a distance consistency reward aimed at minimizing spatial differences between predicted and actual fixation points from the ground truth. The use of these two reward functions can effectively improve the accuracy of point predictions and ensure the personalization of gaze predictions, thereby more effectively aligning human attention patterns.

In summary, our contributions are as follows: 
\begin{itemize}
\item We present the SPA-ADV dataset, a comprehensive eye-tracking dataset featuring individual-level fixation annotations. These annotations are influenced by factors such as age and gender, providing valuable support for benchmark studies in personalized saliency modeling.
\item We propose a Multi-Attribute Point-Based Attention MLLMs 
to directly predict fixation points on high-resolution video frames, guided by user attributes and a task-specific learnable token.
\item We introduce the Consistency Group Relative Policy Optimization algorithm, designed to enforce format and spatial consistency and enhance the controllability of personalized gaze prediction.
\end{itemize}

\section{Related Work}
\label{sec:Related}
\subsection{Saliency Prediction Model}
Saliency prediction is a crucial task in computational neuroscience and computer vision, focusing on identifying regions within a visual scene that most likely attract human attention. Initially, models drew inspiration from biological mechanisms and depended on handcrafted features like color, intensity, and orientation contrast but struggled with complex scenes~\cite{color1,color2}. The emergence of CNN-based models \cite{deepgaze,FastSal,EML-NET} marked a significant advancement by enabling more accurate saliency predictions directly from images. These models improved spatial precision and efficiency through attention mechanisms. As research progressed, there was a shift from purely spatial approaches to those incorporating temporal dynamics. LSTM-based models capture the temporal progression of visual attention across video sequences \cite{sam-resnet,temp-sal}, while transformer-based models enhance contextual understanding in sequential and multimodal tasks \cite{sswin,MDS-ViTNet}. TranSalNet \cite{tran} integrates Transformer and CNN architectures to enhance saliency prediction by capturing both local and global features. Additionally, SUM \cite{SUM} introduces the Mamba framework, improving saliency estimation across various domains and datasets. These advancements indicate a major change from low-level feature engineering to high-level semantic understanding in saliency modeling. However, they lack analysis and attention to the differences of personalized saliency representations, which limits their application in cognitive science and human-computer interaction.

\subsection{Individual Differences in Visual Saliency}
Traditional saliency prediction methods generally assume a unified saliency distribution among observers, disregarding individual differences. However, visual selective attention~\cite{selective} reflects personal preferences in prioritizing visual stimuli. These differences are stable and predictable, influencing group-level attention patterns. For instance, when viewing natural images, people tend to fixate initially on the center of mass \cite{mass}, a phenomenon known as central bias \cite{center}. Notably, age further modulates these attention biases, pseudoneglect illustrates a leftward bias in younger individuals that gradually shifts rightward with age \cite{age}. Consequently, individual gaze patterns can diverge from aggregated group-level trends, making it difficult to capture personalized fixation behavior through group-based predictions alone. Initial efforts by Xu et al.\cite{Xu18} incorporated user-specific features into saliency modeling, though the identification of optimal feature combinations remains an open question. Subsequently, Moroto et al.~\cite{Moroto20} employed a multi-task CNN to learn personalized saliency representations and match unseen users to known profiles via attention similarity. However, it showed limited generalization to unseen user types. More recently, F. Strohm et al.\cite{Strohm24} adopted a single-task CNN with user embeddings to incorporate individual traits directly, while Chen et al.\cite{Chen24} introduced observer-specific encoding mechanisms for scanpath prediction. These collective studies demonstrate the feasibility of modeling individual differences in visual attention and underscore the effectiveness of personalized attention modeling in enhancing saliency prediction.

\begin{figure}[t]
  \centering
  \includegraphics[width=\linewidth]{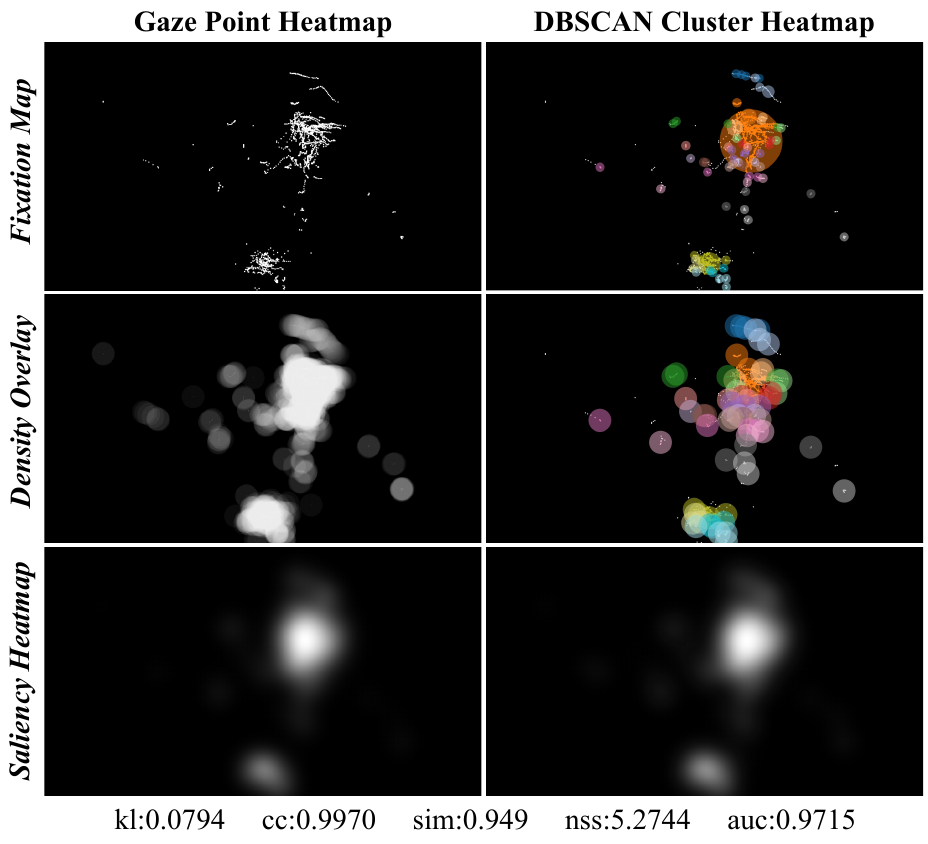}
  \caption{Visualization of gaze point heatmap and DBSCAN clusters heatmap. 
  Fixation heatmap compares raw gaze points with clustered results, where distinct colors denote different clusters. Density Overlay, indicated by uniform-radius circles, demonstrates consistent density distribution between them. The DBSCAN saliency heatmap achieves strong agreement with the original gaze distribution in both visual patterns and quantitative metrics.}
  \label{dataset}
\end{figure}

\subsection{Reinforcement Learning for LLM}
Multimodal Large Language Models (MLLMs) have shown remarkable capabilities in cross-domain problem-solving through integrated processing of visual, linguistic, and spatiotemporal information \cite{gpt,gemini,videollama3,qwen,internvl}. Despite their general effectiveness, MLLMs often struggle to satisfy task-specific objectives or human expectations, particularly in subjective complex output spaces. Reinforcement learning (\textbf{RL}) offers an effective solution by optimizing model behavior through reward-driven feedback. In RL, agents interact with the environment to learn optimal behaviors, with the reward function directing learning by assigning higher values to preferred outputs \cite{RL1,RL2}. When applied to LLMs, RL enables learning beyond supervised data by incorporating richer, often preference-based signals that better reflect task goals or human judgment. Several RL-based optimization strategies have emerged for aligning LLM outputs. Proximal Policy Optimization (PPO) \cite{ppo} performs iterative reward-guided updates with trust-region constraints to ensure stable training. Direct Preference Optimization (DPO) \cite{dpo} improves efficiency by directly learning from preference-labeled data without explicit reward modeling. Building on these methods, Group Relative Policy Optimization (GRPO) \cite{GRPO} introduces a group-wise preference constraint that combines DPO's efficiency with PPO's training stability, showing particular effectiveness in preference alignment scenarios. In summary, by leveraging structured rewards or preference supervision, RL-based methods enable models to internalize complex goals that standard supervised learning struggles to capture. These capabilities establish RL as a foundational methodology for learning multi-attribute feature representations in high-dimensional spaces, driving deeper behavioral optimization in next-generation language intelligence systems.

\section{Dataset}
In this section, we present \textbf{SPA-ADV}, a large-scale eye-tracking dataset designed to investigate cognitive differences in visual attention. This dataset is constructed using advanced eye-tracking technology and records gaze behaviors from a wide range of age and gender groups while watching advertisement videos.

\subsection{Data Collection and Processing}
The dataset consists of 486 advertisement videos spanning multiple industries, such as digital products, finance, food and beverages, FMCG, and beauty. This selection ensures rich visual diversity and contextual variety. We recruited 4,600 participants aged 20-55 years through stratified gender sampling. Each participant viewed between 5 and 20 videos, contributing to a comprehensive and representative sample of gaze behavior. Each video was viewed by 20 to 60 individuals with varied demographics. Prior to data collection, participants underwent a calibration process using an EyeTribe Tracker. Sessions were time-limited and included scheduled breaks to mitigate visual fatigue. Eye-tracking data was sampled at 30 Hz. Gaze coordinates were standardized to the resolutions of either 960×540 or 640×360 to accommodate screen sizeGaze coordinates were standardized to the resolutions of either 960×540 or 640×360 to accommodate screen sizes ranging from 18 to 24 inches and ensure consistency across samples.s ranging from 18 to 24 inches and ensure consistency across samples.

To capture temporal dynamics, we segment gaze points at one-second intervals. As illustrated in Figure \ref{dataset}, human gaze tends to cluster around key visual regions. However, the high density and variability of raw gaze data present challenges for direct modeling. To address this, we apply the Density-Based Spatial Clustering of Applications with Noise (DBSCAN)\cite{DBSCAN} algorithm. This method consolidates fixation points into coherent clusters while filtering out those with minimal attention. The centroid of each cluster is then considered as a refined representation:

\begin{equation}
C = \text{DBSCAN}(P, \epsilon, \text{minPts})
\end{equation}

where $\epsilon$ defines the maximum neighborhood distance, and $\text{minPts}$ denotes the minimum number of points required to form a cluster. To generate a saliency representation, we apply a Gaussian kernel to these centroids, producing heatmaps that closely align with traditional Gaze Point HeatMaps, and serve as effective ground truths, as show in Figure \ref{dataset}. 

\subsection{Data Overview, Tasks and Protocols}
We outline key protocol indicators for our dataset in Table~\ref{tab:data}. The SPA-ADV dataset consists of 486 advertisement videos, with 388 allocated for training and 98 for testing. Given the well-documented distinctions between individual and group level attention patterns~\cite{difference,diffrence2}, the dataset is designed to support analyses of both shared and personalized gaze behaviors in response to dynamic video stimuli.

To facilitate these analyses, we define two experimental protocols. Protocol 1 (P1) measures the aggregated gaze responses across all participants, enabling the exploration of general attention distribution patterns. Protocol 2 (P2) introduces a demographic-aware setting by dividing participants into six subgroups: male, female, males over 30, males under 30, females over 30, and females under 30. To ensure robust fixation representation, we report the average number of fixation points per scene (Avg FixPN): in P1, 523 (train) and 515 (test); in P2, 207 (train) and 197 (test). This setup enables fine-grained analysis of gaze behavior across demographic profiles, supporting personalized attention modeling and audience segmentation in visual media analysis.

\begin{figure*}[t]
  \centering
  \includegraphics[width=\linewidth]{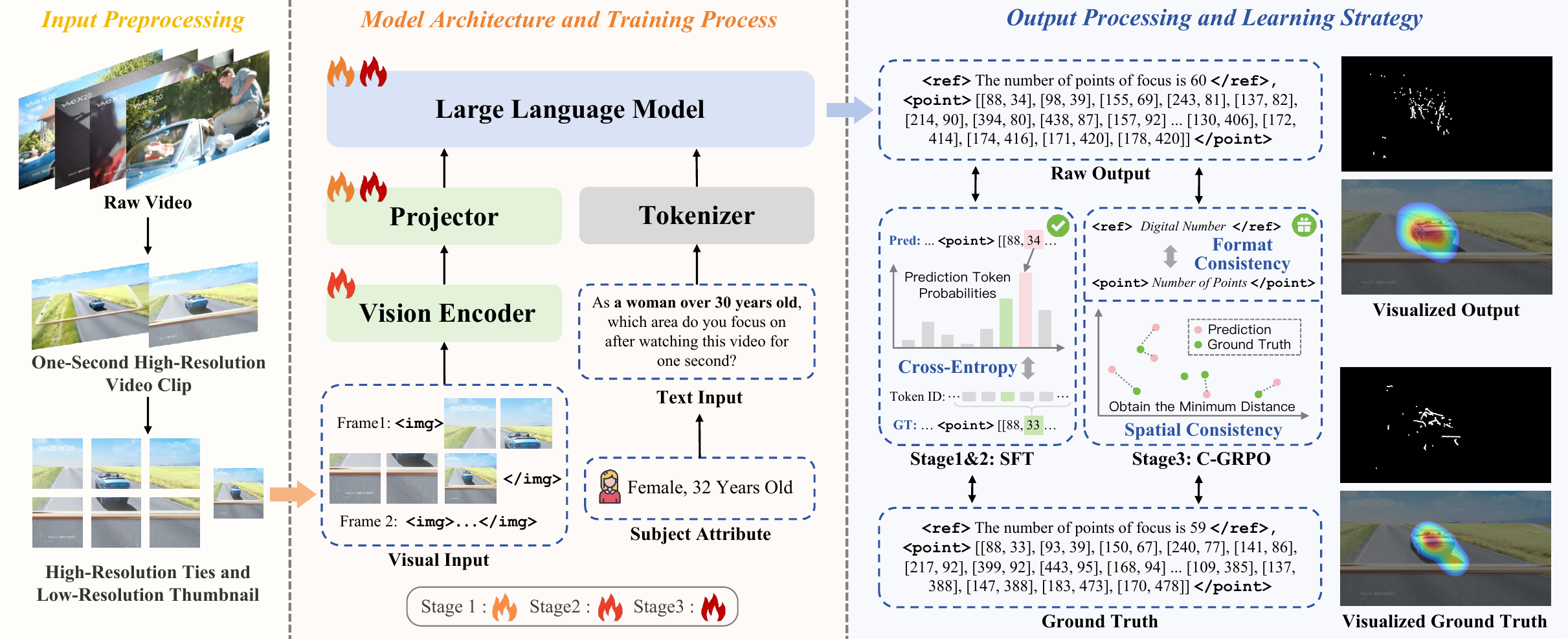}
  \caption{Overview of our PRE-MAP framework. 
  The raw video is divided into one-second high-resolution clips as visual inputs, while subject attributes are organized as textual prompts for Tokenizer. For Stage 1-2 of training process, supervised fine-tuning (SFT) with cross-entropy loss over token IDs is performed using distinct frozen modules. We employs reinforcement learning with C-GRPO in Stage 3 to enhance format and spatial consistency, improving the stability and accuracy of point predictions. The final predicted fixation points are represented as saliency heatmaps for visualization.
  }
  \label{architecture}
\end{figure*}

\begin{table}[h]
    \centering    
    \caption{Sample Numbers and Protocol of SPA-ADV Dataset. Avg FixPN denotes the fixation points number per scene.}
    \label{tab:data}
    \begin{tabular}{ccccc}
        \toprule
         & \textbf{Videos} & \textbf{Protocols} &\textbf{Scenes} & \textbf{Avg FixPN}  \\
        \midrule
        \multirow{2}{*}{\textbf{Train}}  & 388 & P1 & 6265 & 523  \\
                                        & 388 & P2 & 31790 & 207  \\
        \midrule
        \multirow{2}{*}{\textbf{Test}}   & 98 & P1 & 1582 & 515   \\
                                         & 98 & P2 & 8319 & 197   \\
        \bottomrule
    \end{tabular}
\end{table}

\section{Method}
In this section, we introduce the proposed PRE-MAP model along with its reinforcement learning framework. We begin with our Multi-Attribute Point-Based Attention Model, describing how fixation point data is used to guide attention toward visually salient regions (Section~\ref{sec:model}). Section~\ref{sec:CGRPO} presents our reinforcement learning algorithm, termed Consistency Group Relative Policy Optimization (C-GRPO), which is designed to improve both structural and spatial consistency in fixation prediction. The full training pipeline and implementation details are provided in Section~\ref{sec:train}.

\subsection{Multi-Attribute Point-Based Attention Model}
\label{sec:model}

High-resolution inputs are essential for simulating the human visual system's ability to perceive local details and track dynamic changes. To ensure MLLM compatibility, we standardize the resolution of one-second video segments, preserving temporal coherence and visual detail consistency across frames. Each frame is resized to match one of 35 predefined aspect ratios~\cite{internvl}, ranging from 1 to 12 blocks, as shown in Figure~\ref{architecture}. The selected ratio preserves the original image's visual integrity by constraining the resized area to no more than twice the original area. To minimize distortion, the image dimensions are scaled by a base dimension of 448 according to the selected ratio.
The fixation point coordinates $(x, y)$ are then adjusted proportionally:

\begin{equation}
(x', y') = \left( \frac{x}{W} \times W', \frac{y}{H} \times H' \right)
\end{equation}

where $W$ and $H$ denote the original width and height of the image, and $W'$ and $H'$ represent the new dimensions after resizing. To facilitate stable training and improve cross-sample consistency, these adjusted coordinates are further normalized to a fixed grid range of $[0, 1000] \times [0, 1000]$.

To enable effective learning of fixation-relevant regions, we formulate the visual attention patterns through both cardinality prediction of fixation points and coordinate regression parameters, while integrating this geometric representation into LLM via task-specific learnable tokens illustrated in Figure~\ref{architecture}. \texttt{<point>} ... \texttt{</point>} 
represents key spatial anchor coordinates and \texttt{<ref>} ... \texttt{</ref>} tokens represents the expected number of fixation points respectively thereby promoting output consistency and reducing ambiguity or hallucination.

By aligning user multi-attributes (e.g., age/gender) with textual prompts and leveraging fixation annotations, the framework enables more controllable and personalized saliency modeling. This strengthened alignment between predicted fixation maps and human gaze patterns ultimately supports high-resolution, attribute-aware visual content processing across diverse scenarios.

\subsection{Consistency Group Relative Policy Optimization}
\label{sec:CGRPO}

Despite the incorporation of specialized tokens and alignment mechanisms, fixation point prediction still presents persistent challenges in maintaining format integrity and spatial accuracy. To address these limitations, we propose the Consistency Group Relative Policy Optimization (\textbf{C-GRPO}) framework through two novel reward functions: the \textit{Format Consistency Reward} and \textit{Spatial Consistency Reward}.

\textbf{Format Consistency Reward}. This reward encourages the model to generate outputs that strictly follow the predefined structural format. Specifically, it verifies numerical consistency between the number of fixation points specified within the \texttt{<ref>} token($N_{\text{ref}}$) and the number of points generated using \texttt{<point>} markers($N_{\text{actual}}$). The reward is formulated as follows:
\begin{equation}
\small
R_{format} = 
\begin{cases}
0.0, & \text{invalid format} \\
r_{\text{base}} + r_{\text{extra}} \times \dfrac{\min(N_{\text{ref}}, N_{\text{actual}})}{\max(N_{\text{ref}}, N_{\text{actual}})}, & \text{otherwise}
\end{cases}
\end{equation}

If any required special tokens are missing, the reward is set to zero. Otherwise, the model receives a base reward $r_{\text{base}}$, along with an additional reward $r_{\text{extra}}$ that scales with the agreement between the declared and predicted point counts. This reward promotes structural consistency and reduces output hallucinations, thereby improving the reliability of fixation predictions. In our implementation, we use $r_{\text{base}} = 0.2$ and $r_{\text{extra}} = 0.8$. Details regarding reward selection are provided in Section~\ref{sec:reward}.

\textbf{Spatial Consistency Reward}. Since predicted fixation points inherently lack semantic correspondence that prevents direct distance minimization between point sets, we propose a dynamic point matching mechanism through optimal the nearest neighbor regression which enables the model to adaptively suppress outlier predictions while pulling geometrically matched pairs into closer alignment. For each predicted point $p_i \in P$, we identify the nearest ground-truth point $t_j \in T$ and compute the corresponding Euclidean distance. The spatial consistency reward is defined as:
\begin{equation}
    R_{distance} = \exp\left(-\frac{1}{d_{\max} |P|} \sum_{i=1}^{|P|} \min_{t_j \in T} \|p_i - t_j\|^2 \right)
\end{equation}
Here, $d_{\max}$ denotes the maximum possible squared distance in the normalized coordinate space, and $|P|$ is the number of predicted fixation points. When predictions are spatially close to the ground truth, the reward approaches 1; it decays exponentially as the spatial error increases. This formulation penalizes large positional discrepancies and incentivizes accurate spatial localization, thereby improving the precision of fixation prediction.

The overall reward for C-GRPO combines structural and spatial components:
\begin{equation} 
R_{\text{C-GRPO}} = R_{format} +  R_{distance} 
\label{c-grpo}
\end{equation}
This composite reward function guides the policy to generate outputs with format consistency and spatial precision, while ensuring robust generalization across varied input conditions. 
By integrating progressive reinforcement learning with Consistency Group Relative Policy Optimization, our model gains a refined understanding of fixation behavior and achieves superior performance in saliency prediction tasks via accurate fixation forecasting.

\subsection{Training}
\label{sec:train}
Our model is trained on the SPA-ADV dataset through a three-stage pipeline, as shown in Figure~\ref{architecture}. The first two stages involve supervised fine-tuning to progressively adapt the model, while the final stage applies reinforcement learning with the proposed C-GRPO algorithm to enhance the performance of PRE-MAP.

\textbf{Stepwise Fine-Tuning Strategy.}
In stage 1, we first finetune-tune both the language model and visual projector to establish fundamental capabilities for fixation prediction. This joint optimization enables the model to comprehend the task objective through textual instructions, and properly utilize the \texttt{<point>} special token for coordinate regression.This representation demonstrate improving saliency detection accuracy through this initial alignment phase.

Building on the established semantic understanding, we subsequently refine the vision encoder while keeping the language model frozen in Stage 2. This design stems from two key observations: First, transformer-based vision backbones naturally prioritize global scene comprehension over localized details. Second, accurate saliency prediction demands heightened sensitivity to fine-grained visual patterns. The dedicated visual tuning phase enhances spatial acuity without compromising the acquired semantic knowledge.

\textbf{C-GRPO.}
Building upon the GRPO framework, we realize Stage 3 with our proposed C-GRPO. GRPO introduces a group-wise policy optimization mechanism with enhanced stability via policy ratio clipping and KL divergence regularization. This formulation improves the model's robustness and efficiency when learning from structured reward signals. By integrating task-specific reward functions, C-GRPO further enhances predictive performance. 

The final policy update follows the clipped surrogate objective of GRPO, formally defined as:

\begin{equation}
\begin{aligned}
& \mathcal{J}_{\text{GRPO}}(\theta) 
= \mathbb{E} \left[ q \sim P(Q),\ \left\{o_i\right\}_{i=1}^{G} \sim \pi_{\theta_{\text{old}}}(O|q) \right] \\
& \frac{1}{G} \sum_{i=1}^{G} \frac{1}{|o_i|} \sum_{t=1}^{|o_i|} \Bigg\{
\min \left[
r_{i,t} \hat{A}_{i,t},\
\text{clip}(r_{i,t}, 1 - \epsilon, 1 + \epsilon)\ \hat{A}_{i,t}
\right]\\ 
&-\beta\ \mathbb{D}_{\mathrm{KL}} \left[ \pi_\theta \middle\| \pi_{\text{ref}} \right]
\Bigg\}
\end{aligned}
\end{equation}

\begin{equation}where, r_{i,t} = \frac{\pi_\theta(o_{i,t} | q, o_{i,<t})}{\pi_{\theta_{\text{old}}}(o_{i,t} | q, o_{i,<t})}
\end{equation}

Here, $r_{i,t}$ denotes the policy ratio between the current policy $\pi_\theta$ and the previous policy $\pi_{\theta_{\text{old}}}$ at timestep $t$. For each question \( q \), GRPO samples a group of outputs \( \{ o_1, o_2, \dots, o_G \} \) from the old policy \( \pi_{\theta_{\text{old}}} \) and then optimizes the policy model. The term $\hat{A}{i,t}$ is the estimated advantage, and $\beta$ regulates the weight of KL divergence penalty, which constrains the updated policy to remain close to the reference policy $\pi_{\text{ref}}$. To further reduce bias in the policy gradient the reference policy, a KL-regularized gradient correction term is introduced, defined as:
\begin{equation}
GC_{GRPO}(q, o, t, \pi_{\theta_{rm}}) = \hat{A}_{i,t} + \beta \left( \frac{\pi_{\text{ref}}(o_{i,t}|o_{i,<t})}{\pi_\theta(o_{i,t}|o_{i,<t})} - 1 \right) 
\label{eq:kl}
\end{equation}
C-GRPO serves as a stable and generalizable framework for enforcing both output format and spatial consistency. Additional training details are provided in Section~\ref{sec:exp}.

\begin{table}
\small 
\centering
\caption{Comparison of saliency prediction performance on the SPA-ADV dataset. PRE-MAP is evaluated against traditional saliency models (TranSalNet and SUM) and recent MLLMs. 
Bold indicates the best result.}
\label{tab:saliency}
\setlength{\tabcolsep}{3pt} 
\resizebox{1\linewidth}{!}{ 
\begin{tabular}{lcccccc}
\toprule
\textbf{\multirow{2}{*}{Method}} & \textbf{\multirow{2}{*}{Protocol}}  & \multicolumn{5}{c}{\textbf{Saliency}} \\ 

       &           & \textbf{KL↓} & \textbf{CC↑} & \textbf{SIM↑} & \textbf{NSS↑} & \textbf{AUC↑}  \\ 
\midrule
\multicolumn{7}{c}{\textbf{Zero-shot for MLLMs}} \\  
\midrule
\multirow{2}{*}{GPT4o} 
& P1  & 1.9423 & 0.4660 & 0.4602 & 1.2842 & 0.7848  \\  
& P2 & 2.2650 & 0.4097 & 0.4028 & 1.2418 & 0.7807  \\
\multirow{2}{*}{Gemini 2.0 Flash}  
& P1  & 1.4726 & 0.3380 & 0.3751 & 0.8629 & 0.7296  \\
& P2  & 1.6373 & 0.3542 & 0.3490 & 1.0027 & 0.7590  \\
\multirow{2}{*}{Llama 4 Scout}  
& P1  & 3.7166 & 0.3331 & 0.3849 & 0.8828 & 0.7238  \\
& P2  & 3.7434 & 0.3019 & 0.3452 & 0.8848 & 0.7258  \\
\multirow{2}{*}{QwenVL2.5-7B}
& P1  & 12.0586 & 0.0999 & 0.2154 & 0.2578 & 0.5852  \\ 
& P2  & 12.7596 & 0.0762 & 0.1855 & 0.2195 & 0.5753  \\
\multirow{2}{*}{InternVL2.5-8B} 
& P1  & 12.6480 & 0.0572 & 0.1895 & 0.1140 & 0.5769  \\
& P2  & 12.1385 & 0.0604 & 0.1819 & 0.1395 & 0.5859  \\
\midrule
\multicolumn{7}{c}{\textbf{Fine-tune}} \\  
\midrule
Transalnet
& P1  & 0.6428 & 0.7106 & 0.5901 & 2.1867 & 0.8863 \\  
SUM         
& P1  & 1.3418 & 0.7121 & 0.6045 & \textbf{2.2634} & 0.8836 \\  
\midrule
\multirow{2}{*}{InternVL2.5-8B} 
& P1  & 1.2551 & 0.6514 & 0.5640 & 1.7896 & 0.8470  \\ 
& P2  & 2.2496 & 0.5359 & 0.4793 & 1.6439 & 0.8221  \\
\multirow{2}{*}{\textbf{PRE-MAP}}   
& P1 & \textbf{0.6128} & \textbf{0.7648}  & \textbf{0.7010}  & 2.1532 & \textbf{0.8918 } \\ 
& P2 & \textbf{1.5027} & \textbf{0.6698}  & \textbf{0.6253}  & \textbf{2.0595} & \textbf{0.8601}   \\  
\bottomrule
\end{tabular}
}
\end{table}

\section{Experiment}
\label{sec:exp}
\textbf{Evaluation Metrics.} For the saliency prediction task, we conduct evaluations under two experimental protocols, referred to as Protocol 1 (\textbf{P1}) and Protocol 2 (\textbf{P2}). To provide a comprehensive quantitative assessment, we adopt five commonly used metrics: Correlation Coefficient (CC), Similarity Metric (SIM), Kullback-Leibler Divergence (KL), Normalized Scanpath Saliency (NSS), and Area Under the ROC Curve (AUC). Definitions and theoretical foundations of these metrics are detailed in prior studies \cite{metric1}. 

\textbf{Implementation Details.} Our method is built upon the pre-trained InternVL2.5-8B model~\cite{internvl}. By default, we train the model using a stepwise fine-tuning strategy for 5 epochs with a batch size of 64. We employ the AdamW optimizer with an initial learning rate of $4 \times 10^{-5}$, following a cosine annealing schedule with a warm-up ratio of 0.1 and a weight decay of 0.05. The experiments are conducted on a cluster of 8 NVIDIA A100 GPUs, each with 80GB of memory. During the reinforcement learning, the model is further trained for 1 epoch. 

\subsection{Comparison with Baseline Models on the SPA-ADV Dataset}
As detailed in Table~\ref{tab:saliency}, we first evaluate our PRE-MAP on the proposed SPA-ADV dataset under P1, comparing it with representative traditional saliency prediction models. This evaluation demonstrates the effectiveness of our method and validates the benchmark and task formulation. Additionally, we conduct zero-shot evaluations using several  Multimodal Large Language Models, including GPT-4o\cite{gpt}, Gemini 2.0 Flash\cite{gemini}, Llama 4 Scout, Qwen-VL 2.5 (7B)\cite{qwen}, InternVL2.5 (8B)\cite{internvl}, and a fine-tuned InternVL-8B. 

For each MLLM, a one-second video clip along with a task-specific prompt is provided to facilitate comprehension and response generation, under both P1 and P2. Similarly, experiments using classic eye-tracking models, such as Transalnet and SUM, were only conducted under P1. These methods rely solely on visual input and utilize one frame per second from the video.

(1) In the zero-shot setting, GPT-4o, Gemini 2.0 Flash and Llama 4 Scout exhibit a strong central bias, yet fail to generate accurate fixation predictions. Smaller-scale models like Qwen-VL and InternVL show limited task comprehension, likely due to current challenges in instruction following and visual grounding in MLLMs. Most MLLMs remain constrained by difficulties in fine-grained visual alignment and generating spatially coherent saliency predictions.

(2) Upon fine-tuning the models with SPA-ADV, InternVL demonstrates a clear improvement, indicating a basic understanding of the task. However, its predictions still lack precision. In contrast, our proposed PRE-MAP outperforms both traditional saliency models and MLLMs across multiple quantitative metrics. Furthermore, under P2, PRE-MAP demonstrates strong generalization ability, effectively capturing both low-level visual cues and high-level cognitive factors that shape human visual attention.

\begin{table}[t]
\centering
\caption{Comparison of Traditional Eye-Tracking Models on the MIT1003 and SalEC Datasets.}
\label{tab:mit}
\resizebox{1\linewidth}{!}{
    \begin{tabular}{llccccc} 
        \toprule
       \textbf{\multirow{2}{*}{Dataset}} & \textbf{\multirow{2}{*}{Method}}  & \multicolumn{5}{c}{\textbf{Saliency}} \\ 

       &           & \textbf{KL↓} & \textbf{CC↑} & \textbf{SIM↑} & \textbf{NSS↑} & \textbf{AUC↑}  \\ 
        \midrule
        \multirow{7}{*}{\textit{MIT1003}\cite{mit1003}}   
        & FastSal\cite{FastSal}          & 1.036 & 0.590 & 0.478 & 2.008 & 0.875 \\
        & SAM-Resnet\cite{sam-resnet}    & 1.247 & 0.746 & 0.597 & 2.752 & 0.902 \\
        & DAV\cite{DAV}                  & 0.753 & 0.699 & 0.566 & 2.574 & 0.897 \\
        & UNISAL\cite{UNISAL}            & 1.014 & 0.734 & 0.597 & 2.759 & 0.902  \\
        & Transalnet\cite{tran}          & 0.660 & 0.722 & 0.592 & 2.631 & 0.903  \\  
        & SUM \cite{SUM}                 & \textbf{0.563 }& 0.768 & 0.630 & 2.839 & \textbf{0.913} \\  
        & \textbf{PRE-MAP}     & 0.658 &  \textbf{0.773} & \textbf{0.699} & \textbf{2.945} & 0.890 \\  
        \midrule
        \multirow{7}{*}{\textit{SalEC}}  
        & SSM\cite{sam-resnet}             &  0.720 & 0.599 & 0.611 & 1.396 &0.830 \\
        & DeepGaze IIE\cite{deepgaze}     & 0.995 & 0.560   & 0.399 & 1.327 & 0.842 \\
        &EML-NET\cite{EML-NET}            & 1.220 & 0.510 &  0.536 &1.232 &0.807 \\
        & Transalnet\cite{tran}          & 0.873 & 0.717  & 0.534 & 1.723 & 0.824  \\
        & Temp-Sal\cite{temp-sal}          & 0.712 & 0.719  & 0.629 & \textbf{1.768} & 0.813  \\  
        & SSwinTransformer\cite{sswin}   & 0.652 & 0.687  & 0.606 & 1.701  & \textbf{0.868 } \\  
        & \textbf{PRE-MAP}            & \textbf{0.632} & \textbf{0.736}  & \textbf{0.719} & 1.523 & 0.826  \\  
        \bottomrule
    \end{tabular}
    }
\end{table}

\begin{table}[t]
\small
\centering
\caption{Effect of stepwise fine-tuning. Stage 1 and 2 utilize SFT for different model components, while Stage 3 applies our C-GRPO method to refine fixation prediction.
}
\label{tab:train}
\resizebox{1\linewidth}{!}{
\begin{tabular}{lcccccc}  
\toprule
\textbf{\multirow{2}{*}{Method}} & \textbf{\multirow{2}{*}{Protocol}}  & \multicolumn{5}{c}{\textbf{Saliency}} \\ 

       &           & \textbf{KL↓} & \textbf{CC↑} & \textbf{SIM↑} & \textbf{NSS↑} & \textbf{AUC↑}  \\ 
\midrule 
\multirow{2}{*}{Stage1} 
& P1 & 0.6378 & 0.7553  & 0.6819  & 1.8053 & 0.8729  \\  
& P2 & 1.5879 & 0.6609  & 0.6067 & 1.7177 & 0.8525  \\
\multirow{2}{*}{Stage2}   
& P1 & 0.6299 & 0.7648  & 0.6925 & 1.9532 & 0.8863 \\ 
& P2 & 1.5031 & 0.6696  & 0.6224 & 1.7595 & 0.8537 \\
\multirow{2}{*}{Stage3} 
& P1 & \textbf{0.6128} & \textbf{0.7648}  &\textbf{0.7010}  &\textbf{2.1532} & \textbf{0.8918}  \\ 
& P2 & \textbf{1.5027} & \textbf{0.6698}  & \textbf{0.6253}  & \textbf{2.0595} & \textbf{0.8601}   \\
\bottomrule
\end{tabular}
}
\end{table}

\subsection{Comparison on Other Benchmark Datasets }
We evaluate our PRE-MAP on two widely used saliency benchmarks, MIT1003\cite{mit1003} and SalEC, to assess its generalization and robustness, as shown in Table~\ref{tab:mit}. PRE-MAP consistently outperforms baselines across most evaluation metrics, demonstrating its effectiveness in general saliency prediction tasks. While slight decreases are observed in NSS and AUC scores, these are primarily attributed to our decision to retain the original high-resolution input videos without downsampling. Although this preserves fine-grained visual details essential for human-level interpretation, it may also lead to minor spatial misalignments between predicted and ground-truth fixation maps, thereby amplifying pixel-level deviations in saliency prediction.

\begin{figure*}[t]
  \centering
  \includegraphics[width=\linewidth]{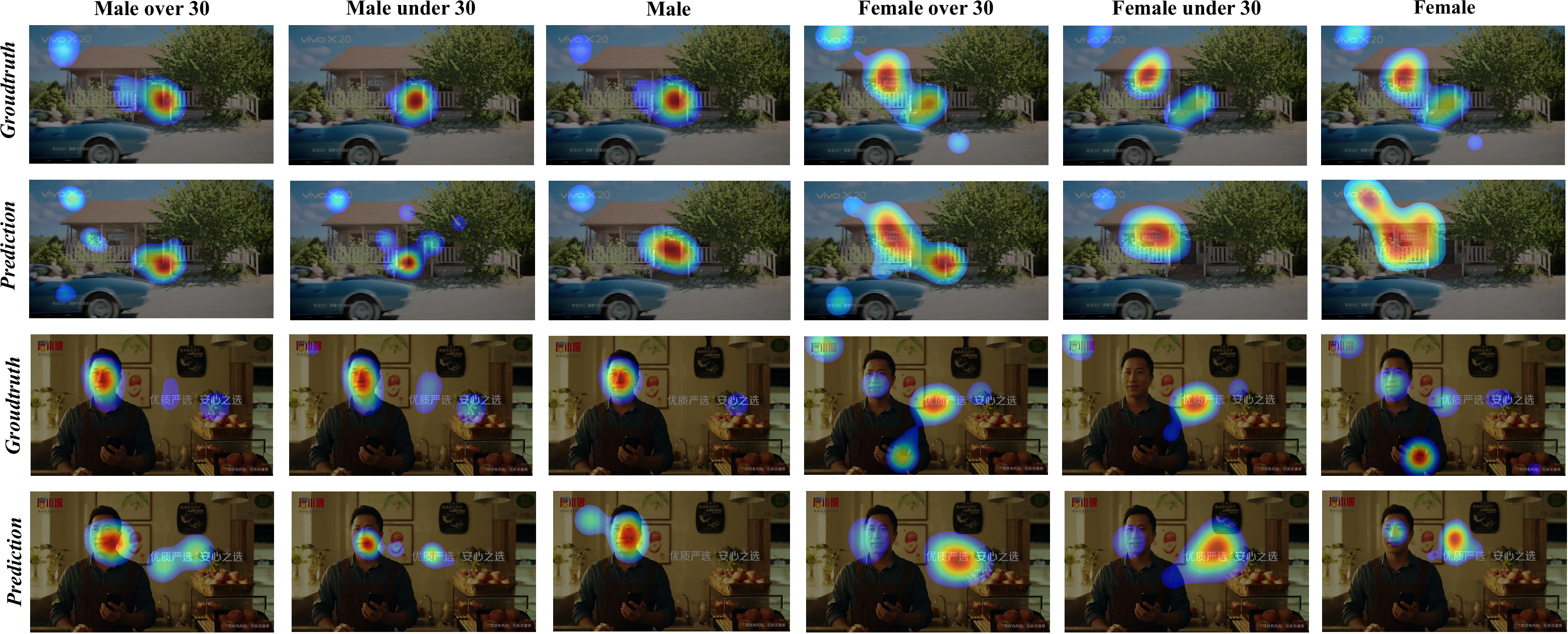}
  \caption{Comparison of saliency heatmaps showing the impact of subjective cognitive processes on visual selective attention under objective visual stimuli, against ground truth.}
  \label{visualization}
\end{figure*}

\subsection{Ablation Study}
To evaluate the contribution of each stage in the training pipeline, we conducted comprehensive ablation studies on the SPA-ADV dataset. We then investigate how DBSCAN parameters affect the alignment between processed saliency maps and ground-truth gaze distributions. Finally, we explore the impact of balancing the format consistency rewards $r_\text{base}$ and $r_\text{extra}$ on model performance, offering insights into the role of reward shaping.

\subsubsection{\textbf{Effect of Stepwise Fine-Tuning and C-GRPO}}
We evaluate the contribution of each stage in the training process, as detailed in Table \ref{tab:train}: (1) fine-tuning the language model alone, (2) further fine-tuning the vision encoder, (3) refining the language model with the proposed C-GRPO algorithm.

Fine-tuning the language model with task-specific tokens used to align subjectively visual grounding outputs with LLM thereby reducing generation ambiguity and improving the model's ability to capture subjective visual attention, which varies due to individual cognitive differences. Further fine-tuning the vision encoder enables the model to better capture fine-grained visual features, resulting in more accurate pixel-level saliency heatmaps. Finally, applying the C-GRPO algorithm refines the spatial precision and structural coherence of predicted fixation, producing saliency maps that more closely align with human-labeled ground truth.

\begin{table}[t]
\centering
\caption{Effect of DBSCAN parameters $\epsilon$ and $\text{minPts}$ during the processing of SPA-ADV. $\text{maxN\_Pts}$ denotes the maximum number of fixation points per image following clustering.}
\label{tab:dbscan}
\resizebox{1\linewidth}{!}{
\begin{tabular}{cccccccc}  
\toprule
\textbf{\multirow{2}{*}{$\epsilon$}} & \textbf{\multirow{2}{*}{\text{minPts}}}  & \multicolumn{5}{c}{\textbf{Saliency}} & \textbf{\multirow{2}{*}{maxN\_Pts}}\\ 

       &           & \textbf{KL↓} & \textbf{CC↑} & \textbf{SIM↑} & \textbf{NSS↑} & \textbf{AUC↑}  \\ 
\midrule 
0.03 & 1 & 0.0873 & 0.9657  & 0.8643 & 3.4035 & 0.9505 & 937 \\  
\textbf{0.04} & \textbf{1} & 0.0992 & 0.9571  & 0.8517 & 3.3684 & 0.9503 & 735  \\  
0.05 & 1 & 0.1140 & 0.9469  & 0.8374 & 3.3198 & 0.9450 & 573 \\  
0.03 & 2 & 0.1739 & 0.9652  & 0.8607  & 3.4149 & 0.9466 & 244 \\
\textbf{0.04} & \textbf{2} & 0.1461 & 0.9585  & 0.8530  & 3.3881 & 0.9428 &  200 \\  
\bottomrule
\end{tabular}
}
\end{table}

\subsubsection{\textbf{Parameter Selection for DBSCAN}}
We conduct an ablation study on the selection of $\epsilon$ and $\text{minPts}$ in DBSCAN to optimize the reduction of fixation points while maintaining the accuracy of the generated saliency maps. Setting $\text{minPts} = 1$ with a small $\epsilon$ retains more fixation points and generates heatmaps that closely resemble the originals. However, after clustering, the maximum number of fixation points per image ($\text{maxN\_Pts}$) often exceeds the token limits of LLMs and increases input complexity. In contrast, a larger $\epsilon$ over-aggregates fixations, leading to overly smoothed and less informative heatmaps. The quantitative effects of these parameters are summarized in Table ~\ref{tab:dbscan}.

We adopt $\epsilon = 0.04$ and $\text{minPts} = 1$ as a balanced configuration. For images with fewer than 100 fixation points, we skip clustering and utilize the raw fixation data directly. If the clustered output exceeds 200 points, a stricter configuration ($\epsilon = 0.04$, $\text{minPts} = 2$) is applied to reduce redundancy. While this configuration may not yield the best performance across all metrics, it consistently achieves an effective trade-off between fidelity and efficiency in both clustering and model input.

\begin{table}[t]
\centering
\caption{$r_\text{{base}}$ and $r_\text{{extra}}$ balance in the Format Consistency Reward under Protocol 1 of SPA-ADV.}

\label{tab:results}
\resizebox{\linewidth}{!}{
\begin{tabular}{ccccccc}  
\toprule
\textbf{\multirow{2}{*}{$r_\text{{base}}$}} & \textbf{\multirow{2}{*}{$r_\text{{extra}}$}}  & \multicolumn{5}{c}{\textbf{Saliency}} \\ 

    &     & \textbf{KL↓} & \textbf{CC↑} & \textbf{SIM↑} & \textbf{NSS↑} & \textbf{AUC↑}\\ 
\midrule 
0.2 & 0.8 & \textbf{0.6128} & \textbf{0.7648} & \textbf{0.7010} & \textbf{2.1532} & \textbf{0.8918}  \\  
0.5 & 0.5 & 0.6231 & 0.7646 & 0.6972 & 2.1478 & 0.8884  \\ 
0.8 & 0.2 & 0.6254 & 0.7647 & 0.6947 & 2.0839 & 0.8867  \\
\bottomrule
\end{tabular}
}
\label{tab:reward}
\end{table}

\subsubsection{\textbf{Effect of $r_\text{{base}}$ and $r_\text{{extra}}$ in the Format Consistency Reward}}
\label{sec:reward}
We examine the impact of balancing base reward ($r_\text{base}$) and the extra reward ($r_\text{extra}$) on model performance within the protocol 1. 
The base reward encourages correct output formatting, while the $r_\text{extra}$ enhances consistency between the number of predicted fixation points in \texttt{<point>} and digital number in \texttt{<ref>}. As shown in Table \ref{tab:reward}, we evaluate three reward configurations. Among these, the configuration $r_\text{base}=0.2, r_\text{extra} = 0.8$ achieves the best overall performance. This indicates that while basic format compliance is necessary, aligning the number and structure of predicted fixation points is crucial for minimizing hallucinated predictions and maintaining the integrity of generated saliency maps.

\subsection{Analysis and Visualization}
We present a visualization of prediction from our PRE-MAP to qualitatively evaluate how subjective cognitive differences affect visual attention, as illustrated in Figure~\ref{visualization}. 

The differences observed in saliency maps across subject groups highlight individual cognitive variations, indicating that attention distribution is influenced by personalized multi-attribute. These visualizations demonstrate that our model effectively captures attribute-dependent gaze patterns. In addition, by generating saliency heatmaps directly from the original high-resolution inputs, our method preserves fine-grained spatial information. This approach not only improves visual accuracy but also prevents common artifacts such as pixelation and blurring, enabling reliable qualitative analysis through high-quality visual representations.

\section{Conclusion}
In conclusion, we present the SPA-ADV dataset and PRE-MAP framework, offering novel insights into subjective cognitive-driven visual attention modeling. Unlike existing benchmarks, SPA-ADV includes real eye-tracking data from a large and diverse demographic, allowing for modeling multi-attribute individual differences in visual attention. PRE-MAP utilizes MLLMs to predict personalized fixation points and generate saliency heatmaps from high-resolution video clips, effectively capturing user-specific gaze behaviors. By integrating C-GRPO, PRE-MAP improves format and spatial consistency, leading to more reliable saliency outputs. Extensive evaluations on the SPA-ADV dataset confirm the effectiveness of our proposed approach.

In future work, we aim to incorporate additional supervision rewards, such as KL divergence and similarity metrics, into C-GRPO to better bridge the gap between fixation points and heatmaps. Additionally, we plan to explore a wider range of audience-specific multi-attribute to further enhance personalization. 

\section*{Acknowledgments}
This work was supported by the Brain-like General Vision Model
and Applications project (Grant No. 2022ZD0160402), Jilin Scientific and Technological Development Program, China (Grant No. 20250102221JC).

\bibliographystyle{named}
\bibliography{ijcai25}

\begin{figure*}[h]
  \centering
  \includegraphics[width=\linewidth]{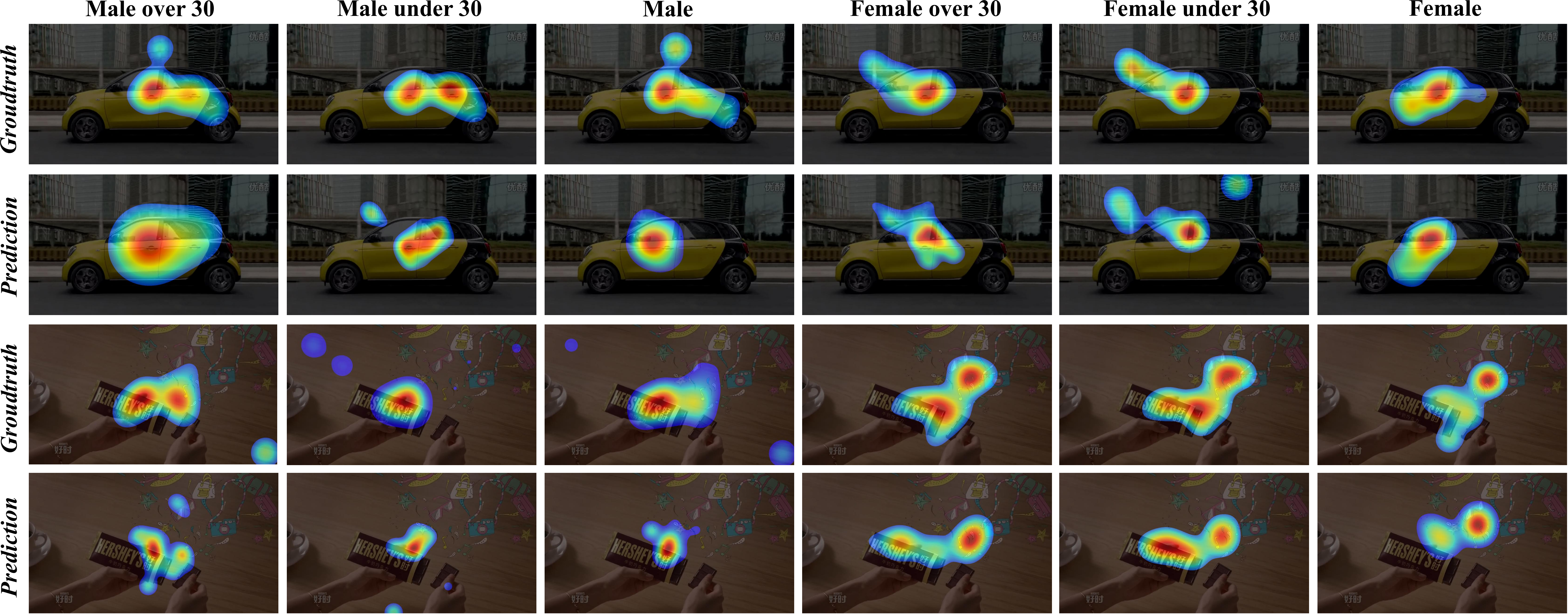}
  \caption{Visualization of saliency maps produced by our PRE-MAP model in comparison with ground-truth maps, highlighting
the impact of subjective cognitive processes on visual attention prediction}
  \label{app}
\end{figure*}

\begin{figure*}[h]
  \centering
  \includegraphics[width=\linewidth]{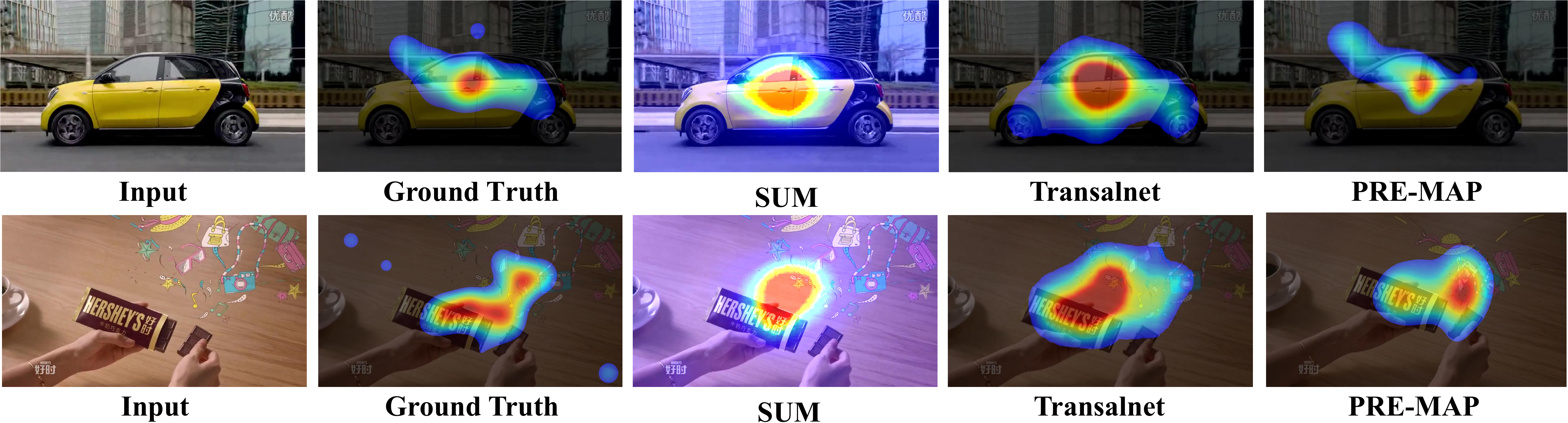}
  \caption{Comparison of saliency maps generated by traditional saliency models with our PRE-MAP, visualized alongside the
ground-truth saliency.}
  \label{app2}
\end{figure*}

\newpage
\appendix
\section{Appendix}
\subsection{Visualization}
Figure \ref{app} presents additional saliency prediction heatmaps generated by our proposed method, PRE-MAP. Figure \ref{app2} shows visualizations of heatmaps produced by traditional saliency models, such as SUM and TransalNet, which downsample the resolution of the original input. For instance, SUM only supports square resolutions of 256 pixels for both input and output, which leads to artifacts in the visualization. Consequently, these conventional models tend to generate highly pixelated heatmaps that primarily emphasize low-level visual features, often resulting in broad and less precise regions of saliency. In contrast, our method (shown in the last column of the Figure \ref{app2} supports inputs at their original resolution and produces more refined predictions by incorporating subjective cognitive processes. This approach effectively captures how individual differences (such as age and gender) influence visual attention.

\subsection{Prompt for MLLMs}
Table \ref{tab:app} presents the complete prompts designed for different MLLMs in our zero-shot experiments on the SPA-ADV dataset, as part of our comparison with baseline models. The full prompt shown for GPT-4o serves as an illustrative example, while prompts for other
models follow the same structure, with variations only in task-specific instructions such as the ’\#\#important note’ to align with model-specific characteristics.

\begin{table*}[t]
    \centering    
    \caption{Model-Specific Prompt Engineering for Zero-Shot Evaluation of Multimodal Large Language Models.}
    \label{tab:app}
    \begin{tabular}{c p{15cm}}  
        \toprule
         \textbf{Model} & \textbf{Prompt} \\
        \midrule
       GPT4o & \textit{As an AI designed for predicting eye movement focus points, evaluate video content to forecast user gaze positions. Consider video details such as themes, structure, colors, and visuals, as well as viewer attributes like age, gender, preferences, and experiences to identify potential focus areas. This involves analyzing how visual elements capture or shift attention, following principles of visual perception and human-computer interaction, focusing solely on prediction results without detailed reasoning.}

\textit{\textbf{\#\# Analysis Factors:}}

\textit{1. Visually prominent elements, such as changes in color and shape.}

\textit{2. Motion and dynamic changes within frames, especially in the final frame.}

\textit{3. Structural complexity and element density in the scene.}

\textit{4. Brightness and light/shadow variations in the image.}

\textit{5. Viewer visual habits and attention concentration patterns.}

\textit{\textbf{\#\# Focus Point Constraints:}}

\textit{1. Quantity: Predict between 1 and 200 gaze points, all as integer coordinates.}

\textit{2. Coordinates: Each focus point is represented as [x, y], within the image’s width and height, 
corresponding to pixel positions.}

\textit{3. Note: Although the output consists of discrete points, each point represents a small region where the viewer is likely to focus, not a single pixel.}

\textit{\textbf{\#\# Important Note:} }

\textit{Always analyze standalone images without assuming privacy or commercial issues. Use existing data and probabilities to provide results, avoiding responses like "I'm sorry, but I can't provide an analysis based on the given images alone."}

\textit{\textbf{\#\# Provide predictions using the following format:}}

\textit{Focus points count is [number], [[x1, y1], [x2, y2], [x3, y3], ...]}

\textit{For example: Focus points count is 62, [[475, 142], [361, 156], [349, 175], [218, 175], [374, 181], [194, 188], [155, 200], ...]}
\textit{
Ensure your predictions follow this format, focusing on precise coordinates only, without additional explanation.} \\
\midrule
Gemini 2.0 Flash & \textit{\textbf{\#\# Important Note:}  }

\textit{It is essential to provide results using existing data and probabilities, and to avoid responses like "Okay, I understand. Please provide the video content and the question." or any similar wording. }\\
\midrule
Llama 4 Scout & \textit{\textbf{\#\# Important Note:} }

\textit{Focus solely on producing output in the specified format below. Do not include any additional text, explanation, or commentary. }\\
\midrule
QwenVL2.5-7B & \textit{No important notes necessary} \\
\midrule
InternVL2.5-8B & \textit{\textbf{\#\# Important Note:} }

\textit{Provide only the integer ratings for the indicators below. Do not include any context, reasoning, descriptions, any other text beyond the ratings, or any Chinese characters.} \\
\bottomrule
\end{tabular}
\end{table*}

\end{document}